\documentclass[runningheads]{llncs}
\usepackage[T1]{fontenc}
\usepackage{amsmath}
\usepackage{amssymb}
\usepackage{subcaption}
\usepackage{wrapfig}
\usepackage{graphicx}
\begin{document}

\title{Foundation Model for Lossy Compression of Spatiotemporal Scientific Data}
\author{
Xiao Li, Jaemoon Lee, Anand Rangarajan, and Sanjay Ranka
}
\authorrunning{X. Li, et al.}
\institute{\begin{tabular}{c}
            \textit{Department of Computer and Information Science and Engineering }\\
            \textit{University of Florida, Gainesville, FL 32611}\\
            %\text{\{xiao.li, j.lee1\}@ufl.edu}, \text{\{anand, ranka\}@cise.ufl.edu}
        \end{tabular}}
\maketitle              % typeset the header of the contribution
\begin{abstract}
We present a foundation model (FM) for lossy scientific data compression, combining a variational autoencoder (VAE) with a hyper-prior structure and a super-resolution (SR) module. The VAE framework uses hyper-priors to model latent space dependencies, enhancing compression efficiency. The SR module refines low-resolution representations into high-resolution outputs, improving reconstruction quality. By alternating between 2D and 3D convolutions, the model efficiently captures spatiotemporal correlations in scientific data while maintaining low computational cost. Experimental results demonstrate that the FM generalizes well to unseen domains and varying data shapes, achieving up to 4$\times$ higher compression ratios than state-of-the-art methods after domain-specific fine-tuning. The SR module improves compression ratio by 30$\%$ compared to simple upsampling techniques. This approach significantly reduces storage and transmission costs for large-scale scientific simulations while preserving data integrity and fidelity.

\end{abstract}

\section{Introduction}

Foundation models (FMs) have attracted significant attention across various domains due to their remarkable versatility and adaptability. In recent years, models like GPT \cite{achiam2023gpt} and SAM \cite{sam} have been proposed to tackle a wide range of tasks, from natural language processing to image task. At their core, these models leverage transfer learning, which allows them to be pre-trained on large, diverse datasets and then fine-tuned for specific tasks.
This paper develops a FM for scientific data compression, an area where these models have largely been unexplored.
% This approach not only enhances their performance on specialized applications but also makes them highly adaptable to new domains with minimal task-specific data.
 Scientific simulations often generate vast amounts of data per run, especially on high-performance computing systems, where the sheer volume and complexity of data present a significant challenge for efficient storage and transmission. Current popular lossy compressors for scientific data largely rely on traditional methods such as wavelet transforms, singular value decomposition, or domain-specific algorithms \cite{fox2020stability,sz3}. While these approaches have proven effective, they often struggle to fully capture the intricacies and high-dimensional relationships inherent in complex simulation data. In contrast, machine learning-based methods have demonstrated remarkable success across multiple domains, suggesting that similar approaches could yield significant improvements in scientific data compression. Given the increasing demand for efficient data storage and processing in scientific simulations, there is a pressing need for research into the application of FMs in the scientific domain and is the main focus of this paper. Foundational models for lossy scientific data compression have to address the following key challenges:
\begin{itemize}
    \item Varying Physics and Spatiotemporal Correlations: Scientific simulations involve complex physics with different spatiotemporal correlations that vary across domains. These variations make it difficult for a single FM to effectively capture all patterns.
    \item Generalization Across Domains: Scientific data varies significantly across fields, making it hard for FMs to generalize without extensive domain-specific fine-tuning.

    \item Wide Data Range and Distribution: Unlike images, scientific datasets have a broader range of values and more complex distributions, including outliers, which pose difficulties for model training and compression.

    \item Guaranteed Error Bounds: Lossy compression in scientific data must ensure that errors do not compromise downstream analyses or simulations. Achieving high compression while maintaining guaranteed error bounds on the primary data.

    \item Multi-resolution output: Decompression should allow for variable output sizes with different resolution depending on the requirement of the application and available hardware resources.
    
\end{itemize}

The proposed foundation model (FM) consists of two main components: a variational autoencoder (VAE) with a hyper-prior structure and a super-resolution (SR) module. The VAE framework incorporates a hyper-prior structure to capture dependencies in the latent space, enhancing compression efficiency. The SR module improves reconstruction quality by refining low-resolution representations. By integrating 3D convolutional layers, the model effectively captures spatiotemporal correlations, enabling efficient representation and compression of complex scientific data with temporal dependencies. The contributions of this paper are as follows:
\begin{itemize}
\item We propose a foundation model that combines VAE and super-resolution architectures for lossy scientific data compression, addressing all the described challenges.
\item We demonstrate that adapting the SR module in the decoder improves compression performance by up to $30\%$ compared to a standard upsampling decoder.
\item Experimental results show that the proposed model generalizes well to unseen domains and varying data shapes, achieving up to 4$\times$ higher compression ratios compared to state-of-the-art compressors.
\end{itemize}
\section{Related Work}

Error-bounded lossy compression provides reliable error control, which is crucial for scientific applications. These methods can be broadly classified into traditional approaches and machine learning-based techniques.

\textbf{Traditional Methods:} Transform-based methods include ZFP \cite{fox2020stability}, which partitions data into 4D blocks and applies nearly orthogonal transformations. MGARD \cite{MGARD_2} employs a multigrid approach to transform data into multilevel coefficients, enabling efficient error-bounded compression. Prediction-based methods focus on predicting data points based on adjacent values. SZ \cite{sz3} predicts each point and encodes the residuals, while DPCM \cite{mun2012dpcm} encodes differences between sequential samples. FAZ \cite{liu2023faz} integrates prediction models and wavelets, providing a flexible framework for compression.

\textbf{Machine Learning-Based Methods:} Machine learning-based methods offer dynamic and adaptive compression strategies, improving performance and efficiency. AE-SZ \cite{liu2021exploring} combines SZ with autoencoder-based compression, while SRN-SZ \cite{liu2023srn} integrates machine learning-driven super-resolution with SZ for hierarchical, error-controlled compression. \cite{xiaoli_roi} introduces a region-adaptive compression approach by combining a UNet ROI detector with an autoencoder, and \cite{xiaoli_atten} proposes an attention-based network to effectively capture key information from hyper-blocks. Despite their promise, machine learning-based methods for scientific data compression remain underexplored.

Recent advances in natural image compression, such as Variational Autoencoders (VAEs)\cite{vae_z,minnen2018joint}, have demonstrated significant potential for lossy data compression. VAEs are particularly effective for modeling the distortion-rate trade-off during training, making them popular for lossy compression. SR models\cite{light_sr,hat_sr}, which reconstruct high-resolution data from low-resolution inputs, also show promise for compression by recovering fine details and improving reconstruction quality.  

However, existing machine learning-based methods predominantly focus on 2D image compression using simplistic decoders, which fail to fully capture the spatiotemporal correlations present in scientific datasets. To overcome this limitation, we extend VAE \cite{minnen2018joint} from 2D image compression to spatiotemporal data and incorporate a SR module as the decoder. This approach not only preserves the temporal dynamics inherent in the data but also enhances the reconstruction fidelity, making it more suitable for complex scientific datasets with intricate spatiotemporal patterns.

\section{Methodology}

\subsection{VAE with Hyperprior}

\vspace*{-0.25cm}

When using transform-based compression, the encoder \(\mathcal{E}_x\) computes the latent space \(\boldsymbol{y} = \mathcal{E}_x(\boldsymbol{x})\) from the input data \(\boldsymbol{x}\). To enhance compression efficiency, the latent space \(\boldsymbol{y}\) is quantized by rounding, yielding \(\tilde{\boldsymbol{y}} = \text{Round}(\boldsymbol{y})\), and then further compressed using lossless entropy coding. The decoder \(\mathcal{D}_x\) reconstructs the data as \(\hat{\boldsymbol{x}} = \mathcal{D}_x(\tilde{\boldsymbol{y}})\). Compression performance is evaluated by distortion \(D\), which measures the difference between \(\boldsymbol{x}\) and \(\hat{\boldsymbol{x}}\), and bit-rate \(R\), which represents the number of bits required to encode \(\tilde{\boldsymbol{y}}\).

To optimize latent space compression, \cite{minnen2018joint} models each element \(\boldsymbol{y}_i\) as a Gaussian distribution \(\mathcal{N}(\boldsymbol{\mu}_i, \boldsymbol{\sigma}_i^2)\), with mean \(\boldsymbol{\mu}_i\) and variance \(\boldsymbol{\sigma}_i^2\). A hyperprior autoencoder (AE) captures the latent space distribution using a hyperencoder \(\mathcal{E}_h\) and hyperdecoder \(\mathcal{D}_h\). The hyperlatent space \(\boldsymbol{z} = \mathcal{E}_h(\boldsymbol{y})\) is quantized to \(\tilde{\boldsymbol{z}}\), and \(\mathcal{D}_h(\tilde{\boldsymbol{z}})\) estimates \((\boldsymbol{\mu}, \boldsymbol{\sigma})\) for \(\boldsymbol{y}\). Each quantized element \(\tilde{\boldsymbol{y}}_i\) is then modeled as:
\begin{equation}
    \tilde{\boldsymbol{y}}_i \sim \mathcal{N}(\boldsymbol{\mu}_i, \boldsymbol{\sigma}_i^2) * \mathcal{U}\left( -0.5, 0.5 \right),
\end{equation}
where \(\mathcal{N}(\boldsymbol{\mu}_i, \boldsymbol{\sigma}_i^2)\) denotes the normal distribution with mean \(\boldsymbol{\mu}_i\) and variance \(\boldsymbol{\sigma}_i^2\), and \(\mathcal{U}\left( -0.5, 0.5 \right)\) represents the uniform distribution introduced by the quantization process, with \( * \) denoting convolution. The conditional probability \(p(\tilde{\boldsymbol{y}} \mid \boldsymbol{\mu}, \boldsymbol{\sigma})\) is then given by:
\begin{equation}
    p(\tilde{\boldsymbol{y}} \mid \boldsymbol{\mu}, \boldsymbol{\sigma}) = \prod_i \left( \mathcal{N}(\boldsymbol{\mu}_i, \boldsymbol{\sigma}_i^2) * \mathcal{U}\left( -0.5, 0.5 \right) \right)(\tilde{\boldsymbol{y}}_i).
\end{equation}
Since there is no prior belief about the hyperlatent space, \(\tilde{\boldsymbol{z}}\) is modeled with the non-parametric, fully factorized density model \(p(\tilde{\boldsymbol{z}})\) \cite{vae_z}. Entropy models, such as arithmetic coding \cite{arithmetic}, can losslessly compress \(\tilde{\boldsymbol{y}}\) given the probability model. Therefore, the bit-rate for \(\tilde{\boldsymbol{y}}\) can be expressed as \( R_{\tilde{y}} = \mathbb{E}_{\tilde{\boldsymbol{y}}} \left( -\log_2(p(\tilde{\boldsymbol{y}} \mid \boldsymbol{\mu}, \boldsymbol{\sigma})) \right) \). The bit-rate for \(\tilde{\boldsymbol{z}}\) can be expressed as \( R_{\tilde{z}} = \mathbb{E}_{\tilde{\boldsymbol{z}}} \left( -\log_2(p(\tilde{\boldsymbol{z}} )) \right) \). The total bit-rate can be calculated as $R = R_{\tilde{y}} + R_{\tilde{z}} $.

\subsection{FM Architecture}
\vspace*{-0.25cm}

\begin{figure}[htb] % Positioning options: h - here, t - top, b - bottom, p - page
    \centering
    \includegraphics[width=1\textwidth]{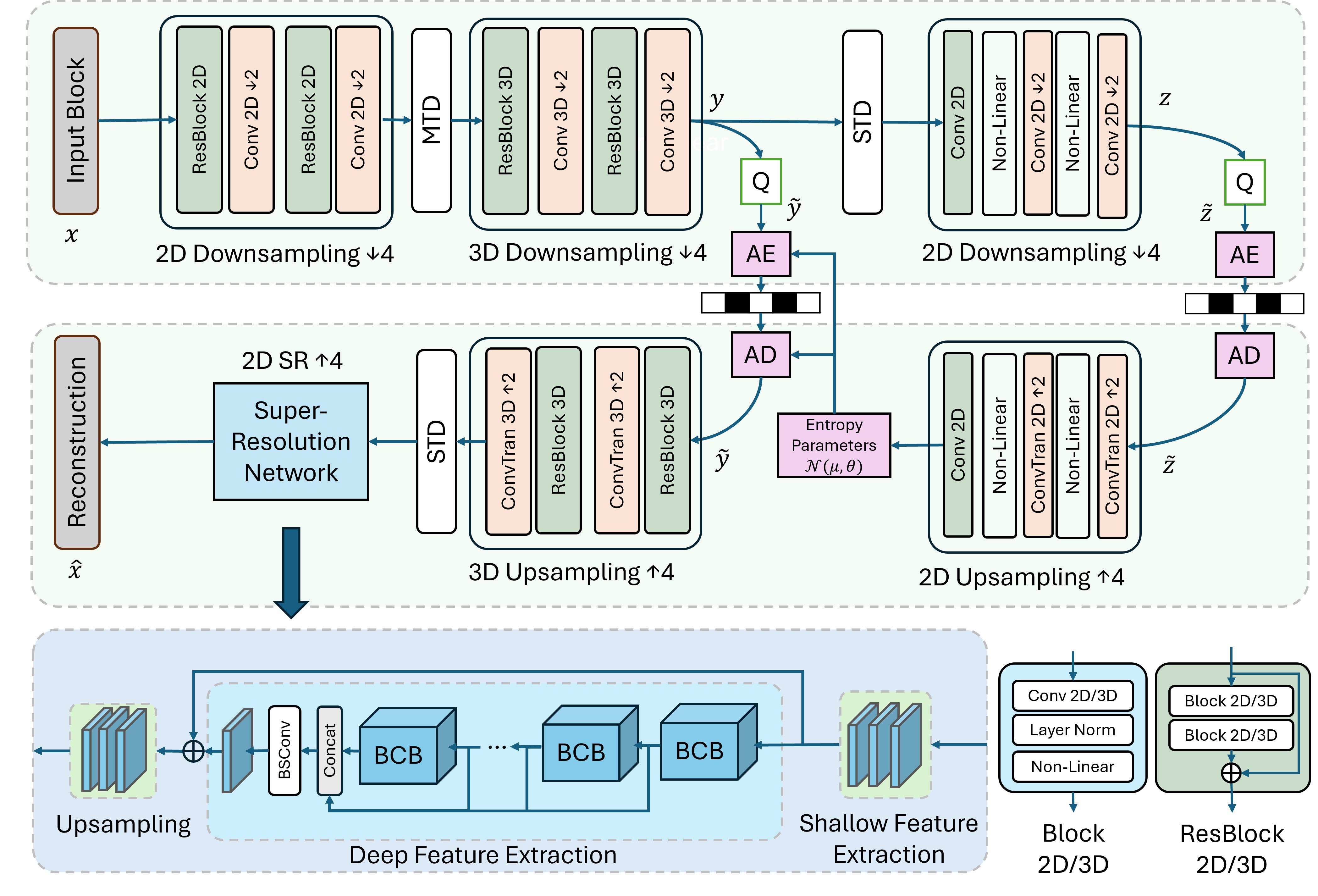} % Adjust width as needed
    \caption{Overview of the architecture of the foundation model (FM). `Conv 2D/3D $\downarrow$' denotes a convolution operation with stride 2, while `ConvTran 2D/3D $\uparrow$' denotes a transposed convolutional layer. `MTD' refers to merging the temporal dimension, and `STD' refers to splitting the temporal dimension. `Q' represents rounding quantization. `AE' and `AD' denote arithmetic encoding and decoding, respectively. Leaky ReLU is used for nonlinearity.}
    \vspace*{-0.35cm}
    \label{fig:overview}
\end{figure}

The FM consists of two main components: VAE with a hyperprior structure and a SR module. First, we extend the hyperprior-based VAE from 2D to 3D to effectively capture spatio-temporal correlations in the data. Next, a 2D SR module is applied to enhance the fidelity of each time slice. These components are described in detail below.
\vspace*{-0.5cm}
\subsubsection{Variational 3D Autoencoder}

We construct a variational 3D autoencoder combining 2D and 3D convolutions to balance efficiency and capacity. While 3D convolutions capture spatio-temporal correlations effectively, they are computationally expensive for large inputs. Therefore, we apply downsampling to the spatial dimensions first and capture temporal correlations at lower-dimensional features. For input dimensions \( [T, H, W] \), where \( T \) is temporal frames and \( H, W \) are spatial, we start with 2D convolutions and downsampling on each temporal slice, producing features \( \left[C, \frac{H}{4}, \frac{W}{4}\right] \) per frame, where \( C \) is the channel count. Then, We merge temporal dimensions (MTD) into \( \left[C, T, \frac{H}{4}, \frac{W}{4}\right] \). Next, 3D convolutions with downsampling reduce dimensions to \( \left[2C, \frac{T}{4}, \frac{H}{16}, \frac{W}{16}\right] \), denoted \( \boldsymbol{y} \). This latent representation is quantized and passed to the decoder. Decoding involves transposed 3D convolutions for upsampling back to \( \left[2C, T, \frac{H}{4}, \frac{W}{4}\right] \), followed by an SR module for independent temporal slice upsampling, reconstructing the data to \( [T, H, W] \).

To model the quantized latent space \( \tilde{\boldsymbol{y}} \), we split the temporal dimension (STD) of \( \boldsymbol{y} \) into \( \frac{T}{4} \) slices. Each slice is independently compressed by a hyper-prior autoencoder (AE), resulting in a hyper-latent representation \( \boldsymbol{z} \) with dimensions \( \left[4C, \frac{H}{64}, \frac{W}{64}\right] \). The hyper-prior decoder upsamples the features to match the original shape of \( \boldsymbol{y} \) while doubling the number of channels. Half of these channels model the mean, and the remaining channels model the standard deviation of the latent space.

\vspace*{-0.5cm}
\subsubsection{SR Module}

In lossy data compression, the decoding phase is essentially an upsampling and interpolation process that closely aligns with tasks in SR. Recent advancements in SR \cite{light_sr,hat_sr} emphasize the effectiveness of more sophisticated architectures compared to conventional upsampling layers, as complex network structures capture finer details and enhance output fidelity.

For our decoder, we adopted a SR architecture inspired by recent work in \cite{light_sr}. As shown in Figure \ref{fig:overview}, the design begins with a shallow feature extractor followed by a sequence of BCB blocks, each enhancing the representation by focusing on hierarchical features. Each BCB block consists of stacked components: an efficient BSConv layer \cite{bsconv}, a ConvNeXt block \cite{convnet}, and two channel attention modules, ESA and CCA. Both BSConv and the ConvNeXt block are widely used in lightweight networks due to their computational efficiency.

\begin{wrapfigure}{r}{0.6\textwidth} % 'r' for right, 'l' for left
    \vspace*{-0.5cm}
    \centering
    \includegraphics[width=0.6\textwidth]{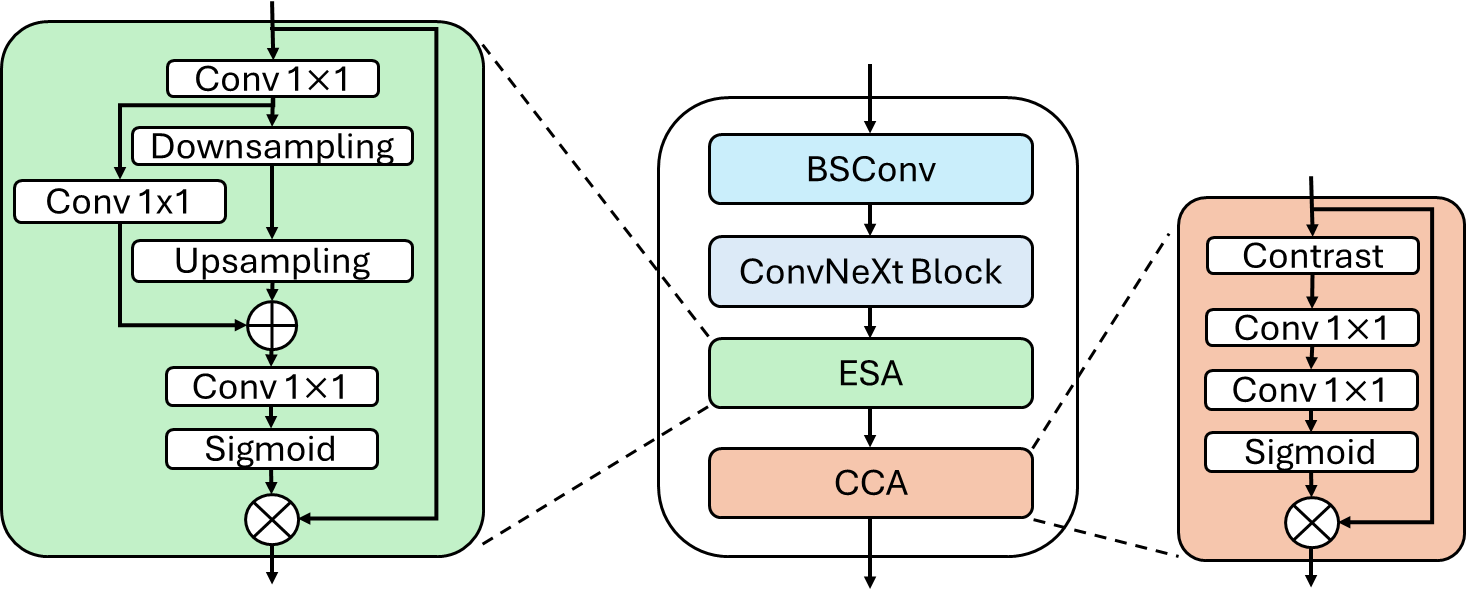} % Adjust width as needed
    \caption{The architecture of the BCB module.}
    \label{fig:srmodel}
    \vspace*{-0.5cm}
\end{wrapfigure}

The ESA module begins by applying a \(1 \times 1\) convolution to reduce the input features' channel dimensions. It then employs stepwise convolutions and max pooling to downsample the feature map. Interpolation-based upsampling restores the feature map to its original spatial dimensions, followed by another \(1 \times 1\) convolution to recover the channel dimensions. Finally, a sigmoid layer generates the attention mask. The CCA module, originally proposed in \cite{cca_module}, starts by combining the mean and standard deviation to generate contrast information. It applies a \(1 \times 1\) convolution to reduce the channel dimensions, followed by another \(1 \times 1\) convolution to restore the original channel size. A sigmoid activation function is then used to generate the attention mask.

As shown in Figure \ref{fig:overview}, the output of each BCB block is concatenated, forming a progressive feature extraction pipeline that captures both shallow and deep representations. This multi-layer aggregation enables the model to retain broad structural information while preserving intricate details. A skip connection merges shallow and deep feature maps, reinforcing the final output by combining low-level texture details with high-level semantic information. The architecture concludes with a channel shuffle layer, used as the primary upsampling mechanism. By rearranging and expanding feature channels, this layer increases the image resolution by a factor of 4.

% This structured approach to decoding, combining channel attention with both shallow and deep feature extraction, enhances the fidelity of reconstructed images and significantly improves upon traditional interpolation-based upsampling methods.

% \subsubsection{Training Objective}

\subsection{Error Bound Guarantee}

To ensure the reconstruction error meets a user-defined error bound, we apply a block-based method. We begin by splitting both the original and reconstructed data into non-overlapping blocks, aiming to minimize the \(\ell_2\)-norm error for each block, \(\left\|\boldsymbol{x} - \boldsymbol{x}^G\right\|_{2}\), where \(\boldsymbol{x}\) represents the original data, and \(\boldsymbol{x}^G\) represents the error-corrected reconstruction. To achieve this, we use Principal Component Analysis (PCA) on the residuals of the data to extract the principal components, or basis matrix, \(\boldsymbol{U}\), with basis vectors ordered by their eigenvalues. We then project each block's residual onto this basis and select the leading coefficients such that the \(\ell_2\) norm of the corrected residual is within a specified threshold, \(\tau\). The residual coefficients are calculated as:
\(\boldsymbol{c} = \boldsymbol{U}^{T}(\boldsymbol{x} - \boldsymbol{x}^R).\)
The most significant coefficients are chosen to satisfy the error bound \(\tau\). To reduce storage costs, we then compress these selected coefficients, \(\boldsymbol{c}_{s}\), using quantization and entropy coding. The final corrected reconstruction is given by:
\(
\boldsymbol{x}^{G} = \boldsymbol{x}^{R} + \boldsymbol{U}_{s}\boldsymbol{c}_{q},
\) where \(\boldsymbol{c}_{q}\) is the quantized set of selected coefficients, and \(\boldsymbol{U}_{s}\) is the corresponding set of basis vectors. We add coefficients until \(\left\|\boldsymbol{x} - \boldsymbol{x}^{G}\right\|_{2} \leq \tau\).

\section{Experimental Results}
 \vspace*{-0.55cm}
\begin{table}[h!]
\centering
\begin{tabular}{c c c c c c}
\hline
Application & Domain & Fields &  Dimensions & Total Size \\ \hline
S3D \cite{s3d} & Combustion & 40  &$40 \times 50 \times 640 \times 640$ & 6.5 GB\\ 
E3SM \cite{e3sm} & Climate &  5  &  $5 \times 1440 \times 240 \times 1440$ & 10.0 GB\\ 
Hurricane \cite{hurricane} & Weather & 3 & $ 3 \times 48 \times 100 \times 500 \times 500$ & 14.4 GB\\ 
JHTDB \cite{jhtdb} & Turbulence & 3 & $ 3 \times 512 \times 512 \times 512$ & 1.6 GB\\ 
% BlastNet & High-Energy Physics & 15 & $ 15 \times 24 \times 64 \times 256 \times 256$ & 6.0 GB\\ 
ERA5 \cite{era5} & Meteorology & 1 & $ 1 \times 4096 \times 300 \times 402$ & 2.0 GB\\ 
Sunquake \cite{sunquake} & Solar Physics & 1 & $ 1 \times  12888\times 256 \times 256$ & 3.4 GB\\ \hline
\end{tabular}
\caption{Datasets Information}
\label{tab:dataset}
\vspace*{-1cm}
\end{table}

 \vspace*{-0.35cm}
\subsubsection{Datasets}
We utilize several datasets to train and evaluate our FM. The E3SM dataset, known for its diverse climate variables and patterns, is specifically used for evaluation, while the other datasets are employed for training.
\textbf{S3D Dataset} \cite{s3d} is a combustion simulation dataset that models the ignition of hydrocarbon fuels under homogeneous charge compression ignition (HCCI) conditions. It captures detailed chemical and thermal dynamics.
% across a $640\times640$ spatial grid over 50 time steps. For this work, we use the mass evolution data of 40 chemical species.
\textbf{E3SM Dataset} \cite{e3sm} is a climate simulation dataset that models various atmospheric and environmental variables. The dataset provides detailed climate dynamics and is used to simulate and predict climate patterns across a wide range of scales. 
% We uses 5 climate variables over 7200 temporal steps, with spatial dimensions of $240 \times 1440$.
\textbf{Hurricane Dataset} \cite{hurricane} simulates the dynamics of tropical cyclones, providing time-varying scalar and vector fields.
% For this work, we use 3 fields over 48 temporal steps, with spatial dimensions of $100 \times 500 \times 500$.
\textbf{JHTDB Dataset} \cite{jhtdb} is a turbulence simulation dataset that provides high-resolution data on fluid dynamics.
% For this work, we use a subset of the JHTDB dataset, focusing on 3 fields over 512 temporal steps, with spatial dimensions of $512 \times 512$.
% \textbf{BlastNet Dataset} is a high-energy physics simulation dataset that models explosive and shockwave phenomena. 
% \\
% For this work, we use a subset of BlastNet, which includes 15 fields with a temporal resolution of 24 steps. The subset consists of 64 sections from the 3D simulation, each with a spatial resolution of \(256 \times 256\).\\ 
\textbf{ERA5 Dataset} \cite{era5} is a global reanalysis dataset that provides detailed information about the Earth's atmosphere, land, and oceans. 
% For this work, we use temperature data consisting of 4096 time slices with a spatial resolution of \(300 \times 402\).\\ 
\textbf{Sunquake Dataset} \cite{sunquake} is an astrophysical dataset that captures seismic-like waves generated by solar flares on the Sun's surface.
% For this work, we use a subset consisting of 12888 temporal slices, each with a spatial resolution of \(256 \times 256\).\\ 
We summarize the details of each dataset in Table \ref{tab:dataset}. The dimensions include the number of fields, the number of temporal slices, and the spatial resolution. For 3D spatial resolution datasets, such as the Hurricane dataset, we use the last two spatial dimensions along with the temporal dimension to construct spatio-temporal blocks. 

\subsubsection{Training Setup}
The training loss consists of two components: distortion \( D \) and bit-rate \( R \). We measure the distortion between the original data \( \boldsymbol{x} \) and the reconstructed data \( \hat{\boldsymbol{x}} \) using the Mean Squared Error (MSE). The bit-rate includes the bit-rate for both the latent space \( \tilde{\boldsymbol{y}} \) and the hyper-latent space \( \tilde{\boldsymbol{z}} \). A factor \( \lambda \) is applied to the bit-rate term to balance the trade-off between distortion and bit-rate. The overall training loss $L$ can be formulated as:
\begin{equation}
    L = \text{MSE}(\boldsymbol{x}, \hat{\boldsymbol{x}}) + \lambda \left( \mathbb{E}_{\tilde{\boldsymbol{y}}} \left[ -\log_2(p(\tilde{\boldsymbol{y}} \mid \boldsymbol{\mu}, \boldsymbol{\sigma})) \right] + \mathbb{E}_{\tilde{\boldsymbol{z}}} \left[ -\log_2(p(\tilde{\boldsymbol{z}})) \right] \right)
\end{equation}

\paragraph{Evaluation Metrics}
We employ the Normalized Root Mean Square Error (NRMSE) as a relative error criterion to evaluate the quality of reconstruction, taking into account that different datasets may span various data ranges. The NRMSE is defined as :
\(
% \label{eq:nrmse}
\mathrm{NRMSE}\left(\Omega,\Omega^{G}\right)=\frac{\sqrt{\left\|\Omega-\Omega^{G}\right\|_{2}^{2}/N_d}}{\max\left(\Omega\right)-\min\left(\Omega\right)},
\)
where $N_d$ is the number of data points in the dataset, and $\Omega$ and $\Omega^{G}$ represent the entire original dataset and the reconstructed datasets after error bound guarantee post-processing, respectively.

\vspace*{-0.25cm}
\paragraph{Configuration}
The FM is trained using PyTorch on two A100-80GB GPUs with the Adam optimizer. Training is divided into two stages: FM training and domain-specific fine-tuning. In the foundation training stage, we run 500,000 iterations, initially setting the weight parameter $\lambda$ to $1 \times 10^{-5}$, which is adjusted to $1 \times 10^{-4}$ after 250,000 iterations. The learning rate starts at $1 \times 10^{-3}$ and is halved every 100,000 iterations. For the domain-specific fine-tuning stage, we conduct an additional 100,000 iterations, setting $\lambda$ at $1 \times 10^{-4}$. Here, the learning rate starts at $1 \times 10^{-4}$ and is reduced by half every 20,000 iterations. The batch size is set to 8 for both stages.
This model is then evaluated on the E3SM application with five climate variables: one for sea level pressure (PSL), two for atmospheric temperature (T200, T500), and two for wind speed (VBOT, UBOT). After obtaining the reconstructed data, we apply an error bound guarantee post-processing step to ensure data quality. Different error bounds are set during post-processing, resulting in varying compression ratios and reconstruction errors.
\vspace*{-0.25cm}

\begin{figure}[htbp]
    \centering
    \begin{subfigure}{0.32\textwidth}
        \includegraphics[width=\linewidth]{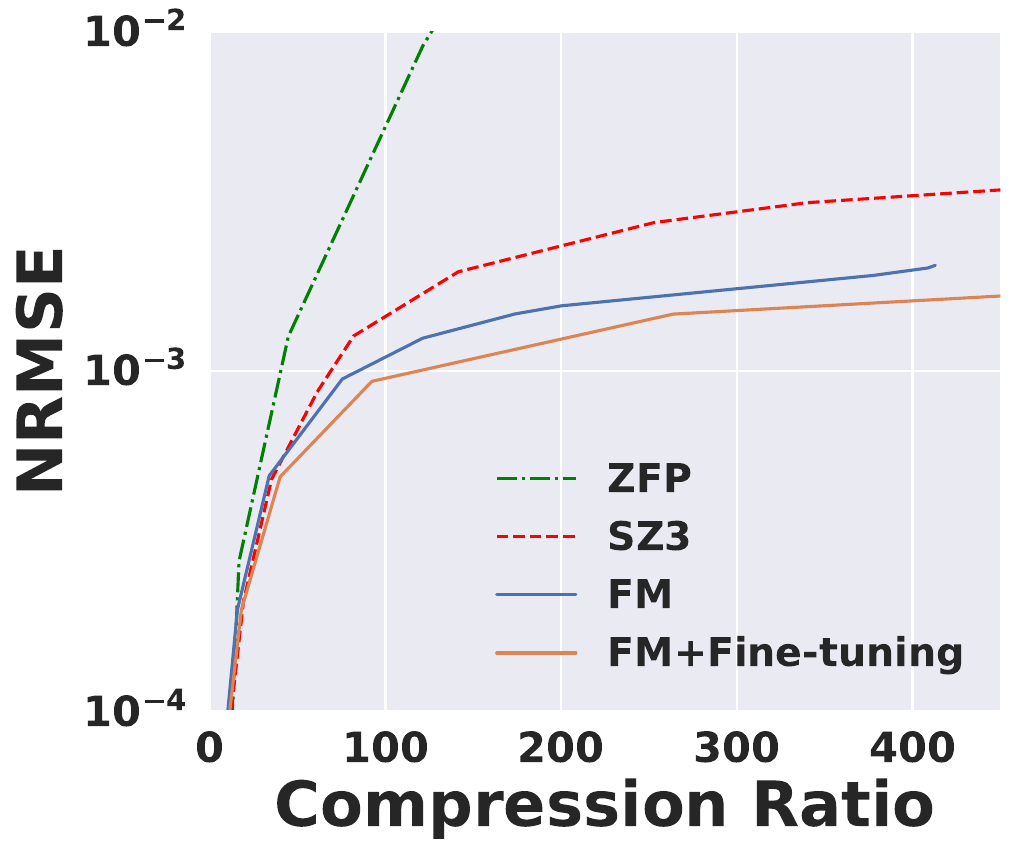}
        \caption{Results on PSL}
        \label{fig:e3sm_fm_var0}
    \end{subfigure}
    \begin{subfigure}{0.32\textwidth}
        \includegraphics[width=\linewidth]{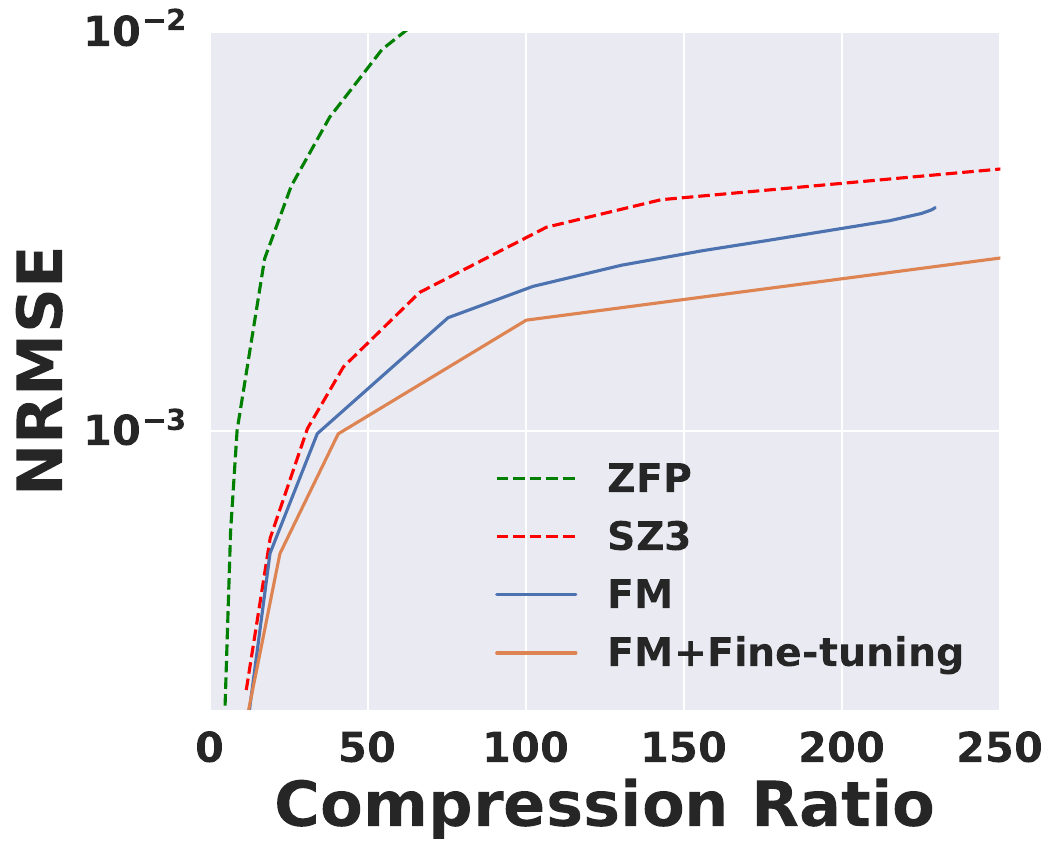}
        \caption{Results on T200}
        \label{fig:e3sm_fm_var1}
    \end{subfigure}
    \begin{subfigure}{0.32\textwidth}
        \includegraphics[width=\linewidth]{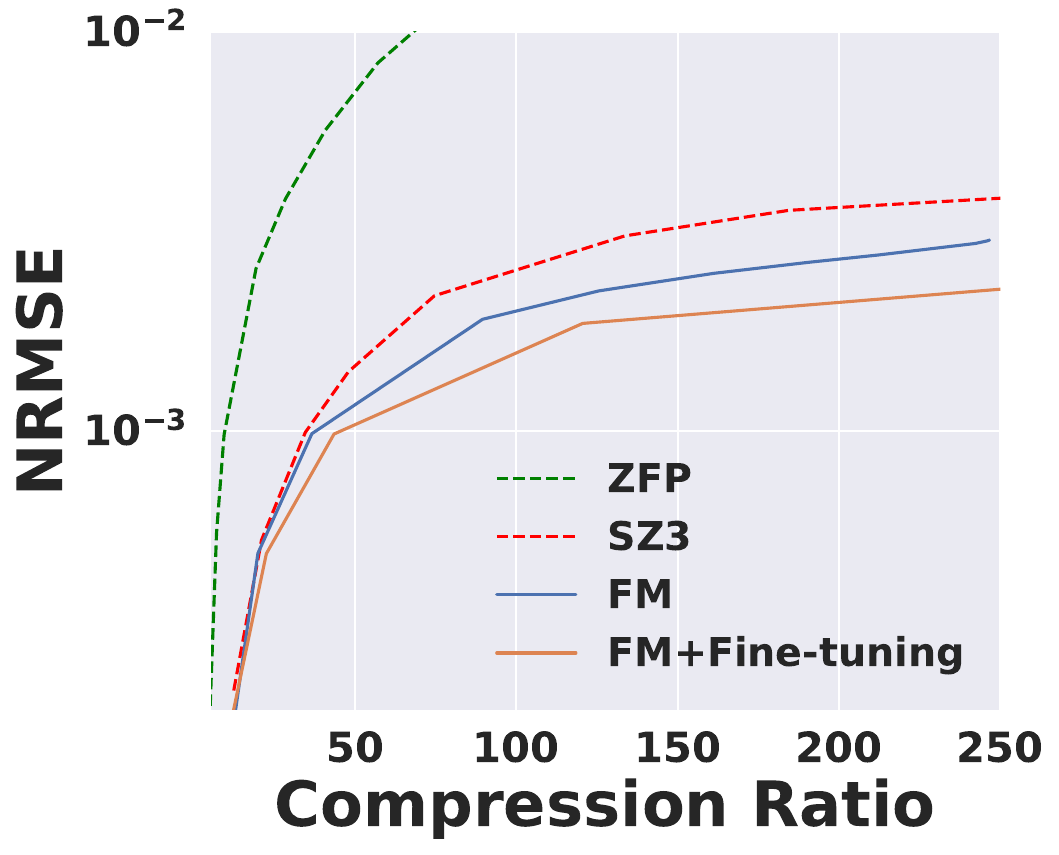}
        \caption{Results on T500}
        \label{fig:e3sm_fm_var2}
    \end{subfigure}
    % \\
    \begin{subfigure}{0.32\textwidth}
        \includegraphics[width=\linewidth]{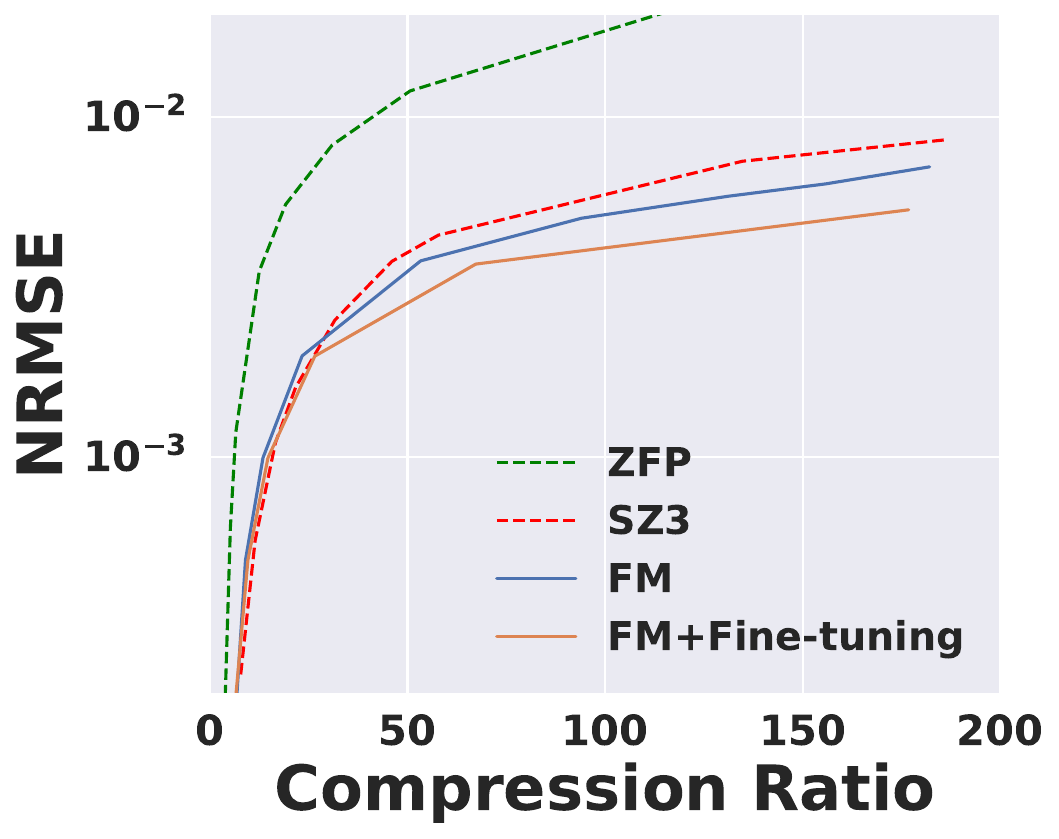}
        \caption{Results on VBOT}
        \label{fig:e3sm_fm_var3}
    \end{subfigure}
    \begin{subfigure}{0.32\textwidth}
        \includegraphics[width=\linewidth]{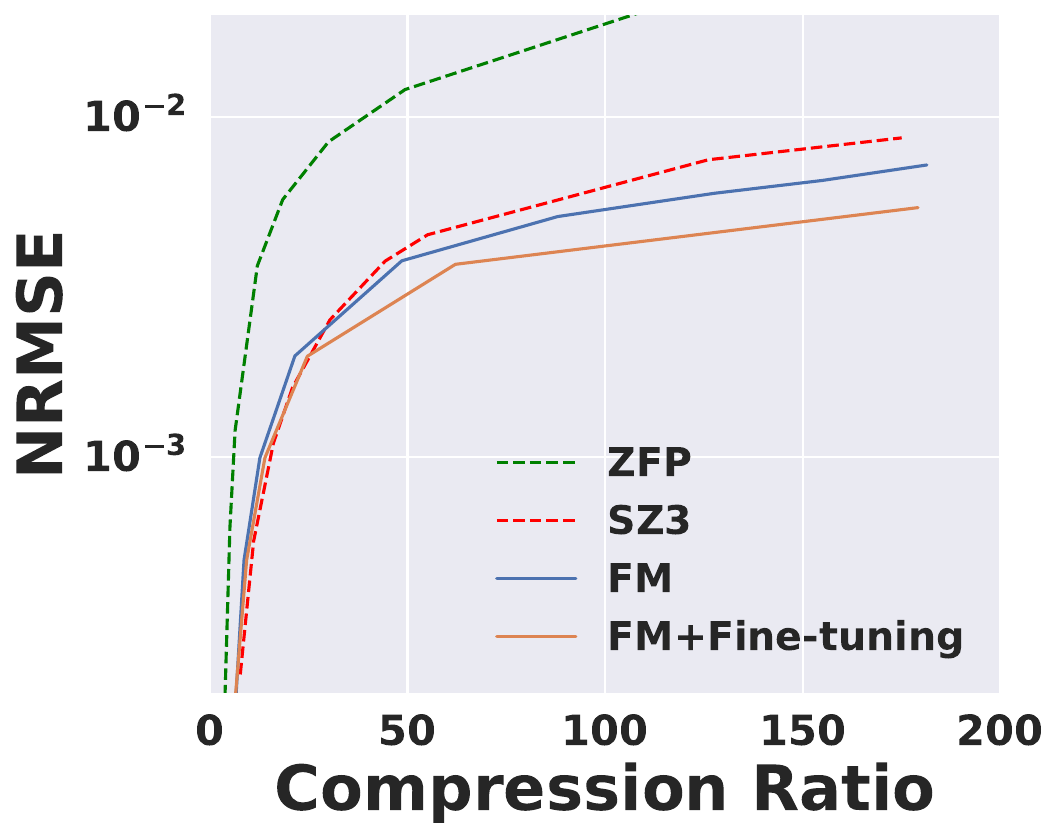}
        \caption{Results on UBOT}
        \label{fig:e3sm_fm_var4}
    \end{subfigure}
    \caption{FM evaluation on 5 variables of E3SM}
    \label{fig:e3sm_fm_vars}
    \vspace*{-0.3cm}
\end{figure}

% \vspace*{-0.5cm}

\paragraph{Data Pre-Processing}
Given that each dataset has a unique data range, typically spanning from $1 \times 10^{-10}$ to $1 \times 10^{10}$, we normalize each field individually within each domain. This normalization is performed by subtracting the mean and dividing by the range of each field, resulting in a normalized data range of $[0, 1]$ with a mean of 0. Our model, based on convolutional layers, is designed to accommodate flexible input shapes and can handle spatial dimensions of $64 N$, where $N \in \mathbb{Z}^{+}$. During training, we prepare data blocks with dimensions $8 \times 256 \times 256$, where 8 represents consecutive time slices, and $256 \times 256$ denotes the spatial resolution. These blocks are generated through random cropping from the full data. For datasets with spatial dimensions smaller than $256 \times 256$, reflection padding is applied to extend each spatial dimension to $256 \times 256$. For evaluation, we maintain an 8-slice temporal dimension and allow flexible spatial dimensions of $64 N$. If an input's spatial dimensions are not divisible by 64, reflection padding is applied to adjust each spatial dimension to the nearest multiple of 64.
To prevent the model from being biased toward larger datasets, we balance dataset sizes by repeating samples from smaller datasets until each dataset has the same number of instances. %This ensures fair representation and consistent learning across all datasets.
\vspace*{-0.25cm}

\subsubsection{Generalization to Unseen Domains}
\begin{figure}[h!]
    \centering
    \includegraphics[width=1\textwidth]{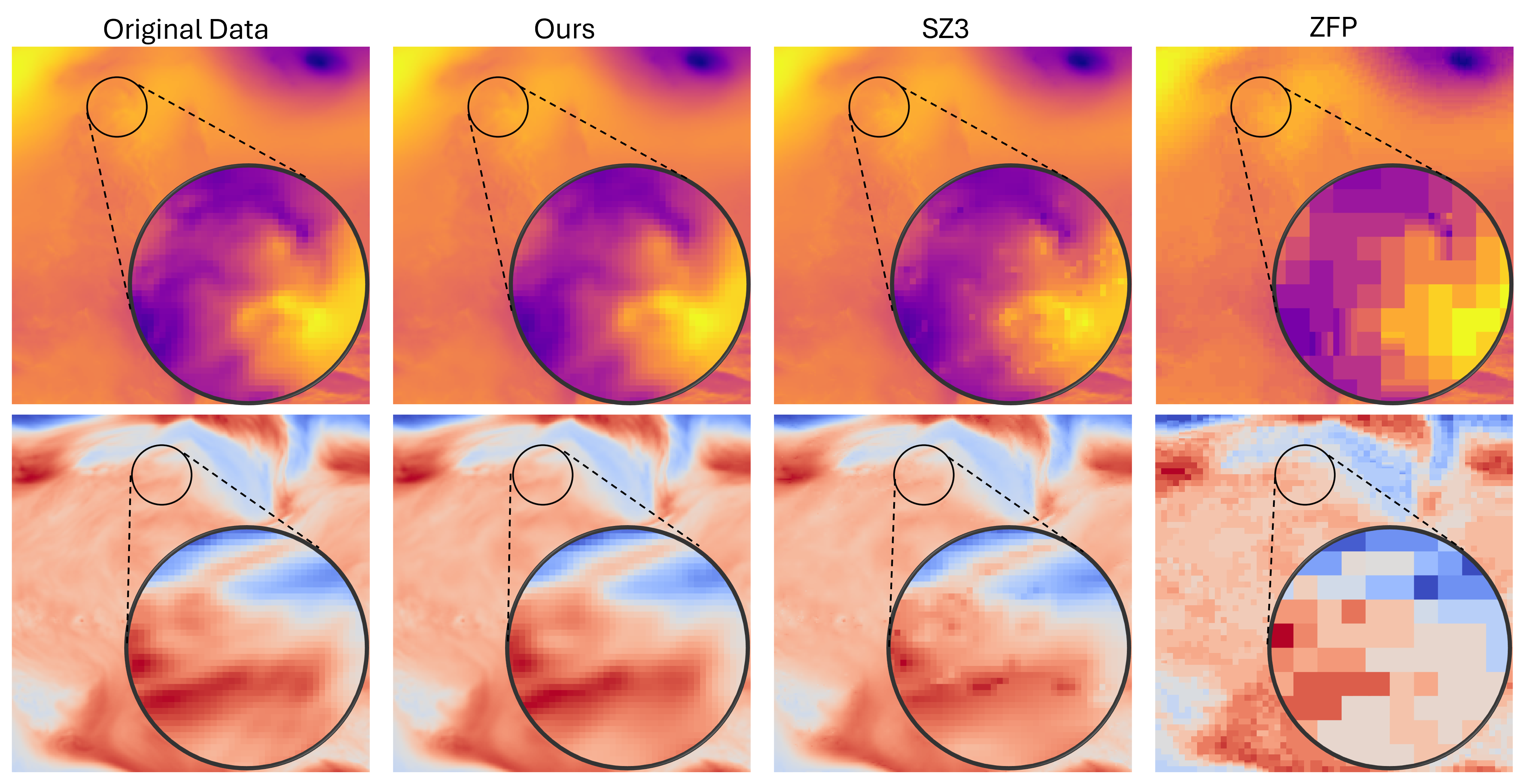} % Replace 'your_graph.png' with your file name
    \caption{Visualization of reconstructed data for our method, SZ3, and ZFP at the compression ratio of 100. The first row shows pressure data (PSL), and the second row shows temperature data (T200).}

    \label{fig:vis} % Optional: For referencing the figure
    \vspace*{-0.7cm}
\end{figure}

To demonstrate the generalization ability of our FM to unseen domains, we evaluate it on the E3SM simulation dataset, which was not used for model training. We compare the performance of our method with two state-of-the-art error-bounded lossy compressors designed specifically for scientific data compression. We present results for our FM both with and without domain-specific fine-tuning, as shown in Figure \ref{fig:e3sm_fm_vars}. Compared to SZ3 \cite{sz3} and ZFP \cite{fox2020stability}, our FM achieves comparable or higher compression ratios at the same reconstruction error level. Domain-specific fine-tuning further boosts compression ratios, providing significantly lower reconstruction error at high compression ratios and similar reconstruction error at low compression ratios. With domain-specific fine-tuning, Our method achieves up to a 4\(\times\) higher compression ratio for the PSL variable and up to a 3\(\times\) higher compression ratio for the two temperature and two wind speed variables, while maintaining the same NRMSE. We also visualize the reconstructed data for our method, SZ3, and ZFP at a compression ratio of 100 in Figure \ref{fig:vis}. Our method shows lower reconstruction distortion compared to other compressors.

\subsubsection{Adaptability to Data Dimensions}
\begin{figure}[h!]
\vspace*{-0.25cm}
    \centering
    \begin{minipage}[b]{0.32\textwidth}
        \centering
        \includegraphics[width=\textwidth]{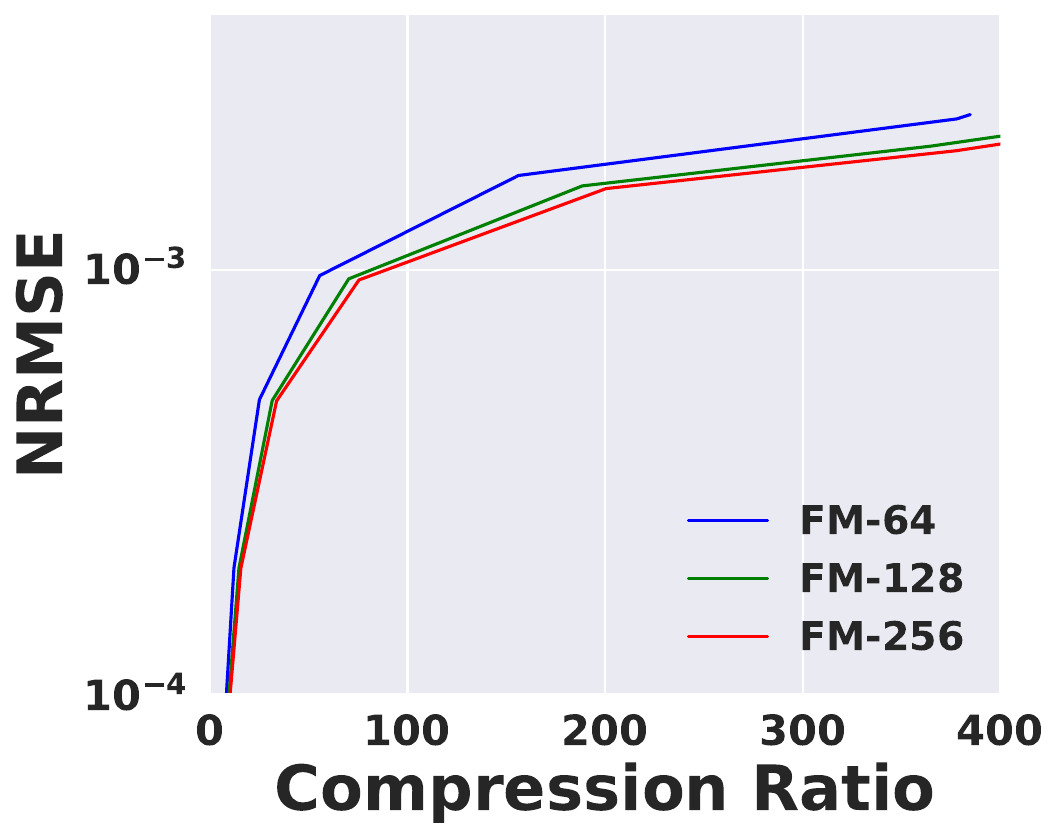}
        \caption{Adaptability to data dimensions}
        \label{fig:resolution}
    \end{minipage}
    \hspace{0.04\textwidth}  % Space between the two images (adjust as needed)
    \begin{minipage}[b]{0.32\textwidth}
        \centering
        \includegraphics[width=\textwidth]{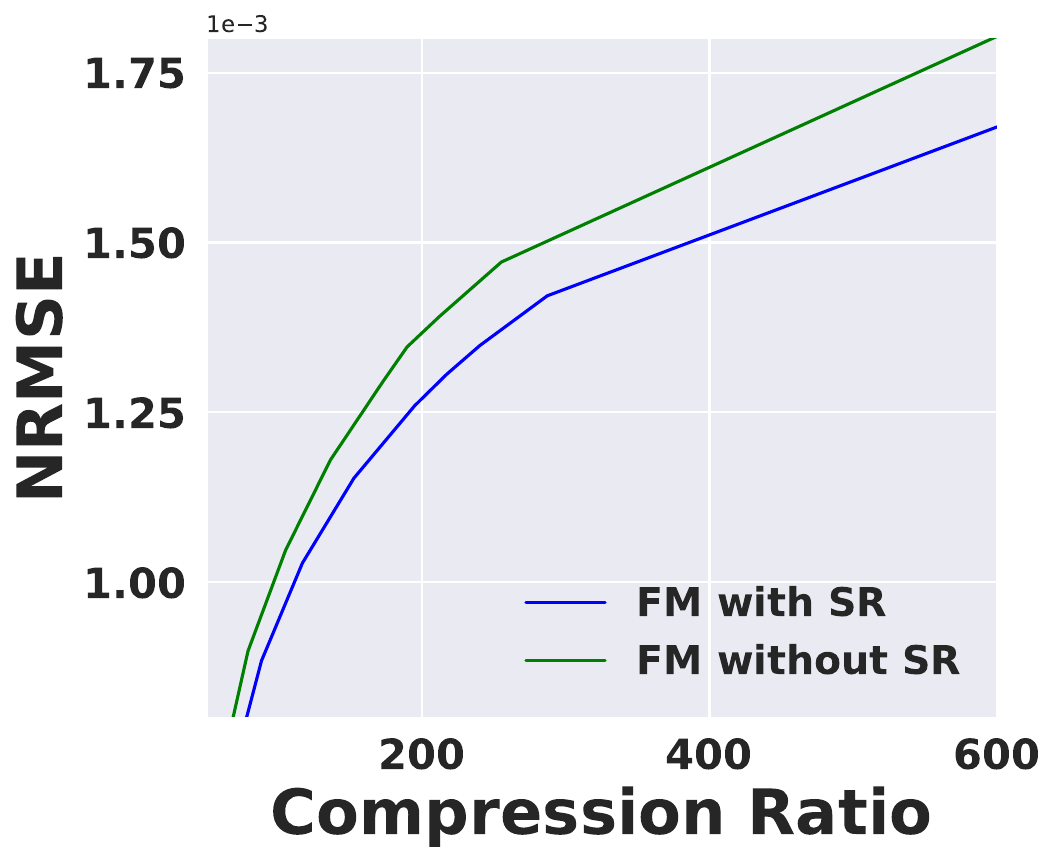}
        \caption{Ablation study for SR module}
        \label{fig:sr}
    \end{minipage}
    \vspace*{-1.2cm}
\end{figure}

Although the model is trained with a spatial size of 256, it supports flexible input shapes and can accommodate spatial dimensions of $64 N$, where $N \in \mathbb{Z}^{+}$. Scientific simulations often generate data with varying spatial dimensions. For using our FM model, data blocking should be applied as a preprocessing step and then the blocked data should be fed into the FM according to their specific requirements. In this experiment, we evaluate the adaptability of our FM model to different spatial resolutions using  E3SM-PSL dataset as the testing dataset. we partition testing data into non-overlapping blocks and feeding the model with spatial resolutions of 64, 128, and 256, while keeping the temporal dimension fixed at 8. The results, presented in Figure \ref{fig:resolution}, demonstrate that our model can effectively handle different block sizes, showing promising compression performance. The result also shows that smaller block sizes generally result in lower compression ratios, as the model captures less structural information within each block. Larger blocks, on the other hand, allow the model to exploit more spatial correlations and patterns, enabling more efficient data representation.

\subsubsection{Impact of Super-resolution}

To evaluate the impact of the SR module on data compression, we compare our FM to a baseline approach where the SR module is replaced with stacked convolutional and non-linear layers. In the baseline approach (refered as `FM without SR'), 2D Transpose Convolution is used as the upsampling operator.  The FM incorporating SR is refered as `FM with SR'. We evaluate these two architectures on E3SM-PSL dataset. The results of these two architectures are presented in Figure \ref{fig:sr}.  As shown in the figure, the SR module significantly improves the compression ratio at the same NRMSE, particularly at high compression levels. For example, at an NRMSE of approximately $1.6 \times 10^{-3}$, the model with SR achieves a compression ratio of 498, compared to 384 for the model without SR---an improvement of roughly 30\%. However, as the compression ratio decreases, the difference between the two models narrows because both employ the same post-processing method to guarantee error bounds. At lower compression levels, the storage cost of the post-processing becomes the dominant factor.  The model size of `FM with SR' is 1.8 million parameters, slightly larger than the 1.5 million parameters in `FM without SR'. Despite this modest increase in size, the substantial improvement in performance demonstrates that the inclusion of the SR module is well justified. This improvement is achieved through more sophisticated hierarchical feature learning, rather than relying on simple upsampling or interpolation methods.

\section{Conclusion}

We present a robust and adaptable foundation model for lossy compression of scientific data, addressing challenges such as domain generalization, spatiotemporal correlations, and error control. By combining a variational autoencoder (VAE) with a hyper-prior structure and a super-resolution (SR) module, the model achieves up to 4 $\times$ higher compression ratios and improved reconstruction quality compared to state-of-the-art methods. Experimental results demonstrate the model's ability to generalize across unseen scientific domains and data configurations, efficiently compressing diverse datasets. The integration of advanced architectures, such as SR networks, further enhances performance and meets the growing demand for efficient scientific data storage and processing. This foundation model represents a significant advancement in data compression, offering improved efficiency and better data integrity preservation.

\bibliographystyle{splncs04}
\bibliography{ref}

\end{document}